\documentclass{article}

\usepackage{algorithm, algpseudocode}
\usepackage[english]{babel}
\usepackage[utf8]{inputenc} 
\usepackage[T1]{fontenc}    
\usepackage[margin=1.25in]{geometry}
\usepackage{graphicx}
\usepackage{hyperref}       
\usepackage{url}            
\usepackage{booktabs}       
\usepackage{amsfonts, amsmath, amssymb}       
\usepackage{nicefrac}       
\usepackage{microtype}      
\usepackage{multirow}
\usepackage{sourcecodepro}
\usepackage{subcaption}

\providecommand{\keywords}[1]{\textbf{\textit{Index terms---}} #1}

\title{Distributed Evolution of Deep Autoencoders}

\author{
  Jeff Hajewski\thanks{Work done while at the Department of Computer Science, University of Iowa} \\
  Salesforce\\
  jhajewski@salesforce.com \\
   \and
  Suely Oliveira \\
  Department of Computer Science\\
  University of Iowa\\
  \and
  Xiaoyu Xing\footnotemark[1]\\
  Amazon\\
}

\begin{document}
\maketitle

\begin{abstract}
  Autoencoders have seen wide success  in domains ranging from feature selection
  to information retrieval.  Despite this success, designing an  autoencoder for a
  given task remains  a challenging undertaking due to the  lack of firm intuition
  on  how the  backing neural  network architectures  of the  encoder and  decoder
  impact the  overall performance of  the autoencoder. In  this work we  present a
  distributed system  that uses  an efficient evolutionary  algorithm to  design a
  modular  autoencoder. We  demonstrate the  effectiveness of  this system  on the
  tasks of manifold  learning and image denoising. The system  beats random search
  by nearly an order of magnitude on both tasks while achieving near linear
  horizontal scaling as additional worker nodes are added to the system.
\end{abstract}


\section{Introduction}
Autoencoders are an unsupervised, deep learning technique used in a wide range
of tasks such as information retrieval (e.g., image search), image restoration,
machine translation, and feature selection. These applications are possible
because the autoencoder learns to distill important information about the input
into an intermediate representation.  One of the challenges in moving between
application domains (e.g., from manifold learning to image denoising) is that it
is not clear how, or even if, one should change the neural network architecture
of the autoencoder.  Designing a neural network is difficult due to the time
required to train the network and the lack of intuition as to how the various
layer types and hyper-parameters will interact with each other. In this work, we
automate the design process through the use of an efficient evolutionary
algorithm coupled with a distributed system that overcomes the computational
barrier of training and evaluating a large number of neural networks.  Our
system is designed to be robust to node failures, which are particularly common
in neural architecture search, and offers elastic compute abilities, allowing
the user to add or remove compute nodes without needing to pause or restart the
search process.

Elastic compute abilities are particularly important in the domain of neural
architecture search due to the high computational demands. For example Zoph et
al. \cite{45826} used 800 GPUs over the course of about a week during their
architecture search of over 12,000 different deep neural networks.  Similarly
\cite{DBLP:conf/icml/RealMSSSTLK17} used 250 GPUs for over 10 days and
\cite{DBLP:journals/corr/abs-1711-00436} used 200 GPUs for a day and a half. At
these levels of compute nodes and experiment duration, node failure is not
surprising.  Additionally, the required level of computational resources may not
be immediately apparent until later stages of the search where potentially more
complex architectures are being explored; having to restart the experiment with
additional compute nodes, rather than simply adding nodes while the experiment
is running, is costly both in terms of research time and money.

We apply this system to the domains of manifold learning \cite{ma2011manifold,
talwalkar2008large} and image denoising \cite{Vincent:2008:ECR:1390156.1390294,
DBLP:journals/jmlr/VincentLLBM10}, which are two common domains of application
for autoencoders.  We explore the effect of varying the number
of epochs during the evolutionary search model evaluation, the scalability of
the system, and the effectiveness of the search.

The primary contributions of this work are:
\begin{itemize}
\item An efficient and scalable evolutionary algorithm for neural architecture
  search applied to the evolution of deep autoencoders.
\item A distributed architecture for the efficient search and evaluation of
  neural networks.  This architecture is robust to node failures and allows
  additional compute resources to be added to the system while it is running.
\item A demonstration of the effectiveness of both the search algorithm and the
  distributed system used to perform the search against random search.  This
  demonstration is performed in the domains of manifold learning and image
  denoising.
\end{itemize}

The rest of this work is organized as follows. We discuss related and prior work
in Section~\ref{sec:related}. In Section~\ref{sec:autoencoders}
we give a brief overview of autoencoders and formally introduce the applications
of manifold learning and denoising. We follow the background material with a
description of our search algorithm along with the architecture and features of
the distributed system we built to efficiently run our search algorithm.
Section~\ref{sec:experiments} contains a description of our experiments, followed
by a discussion of the results, in Section~\ref{sec:results}.

%

\section{Related Work}\label{sec:related}
Neural architecture search has recently experienced a surge of interest, with a
number of clever and effective techniques to find effective neural network
architectures \cite{pmlr-v80-pham18a, 45826, DBLP:journals/corr/abs-1806-09055,
  DBLP:journals/corr/HaDL16, DBLP:journals/corr/abs-1708-05344}. A number of
recent works have explored the use of reinforcement learning to design network
architectures \cite{Kyriakides:2018:NAS:3200947.3208068, 45826, pmlr-v80-pham18a}.
Other recent work \cite{Koutnik:2013:ELN:2463372.2463509, DBLP:journals/corr/MiikkulainenLMR17,
DBLP:journals/corr/abs-1810-05749} has used a similar strategy to exploring the search space of
network architectures, but use evolutionary algorithms rather than reinforcement
learning algorithms. The reinforcement learning and evolutionary-based approaches,
while different in method of search, use the same technique of building modular
networks---the search algorithm designs a smaller module of layers and this module
is used to assemble a larger network architecture. Liu et al.
\cite{DBLP:journals/corr/abs-1711-00436} use this modular approach at multiple levels
to create motifs within modules and within the network architecture itself.

Despite the wealth of prior work on evolutionary approaches to neural architecture
search, to the best of our knowledge there is relatively little work exploring
the application of these techniques to autoencoders. Most similar to our work
is that of Suganuma et al. \cite{pmlr-v80-suganuma18a}, which uses an evolutionary
algorithm to evolve autoencoder architectures.
We use a parent population of 10 rather than the single parent approach used
by \cite{pmlr-v80-suganuma18a}. We are able to do this efficiently for any
number of parents by horizontally scaling our system.
Perhaps the most striking difference between
our work and Suganuma et al. is the improved efficiency we achieve by caching
previously seen genotypes along with their fitness and thus reducing the computational
load by avoid duplicate network evaluations. Additionally, our approach searches for
network architectures asynchronously and on a much larger scale using a distributed
system. The graphical approach used by Suganuma et al. allows them to evolve
non-sequential networks while our work only considers sequential networks -- this
is a shortcoming of our approach.
Lander and Shang
\cite{7273701} evolve autoencoders with a single hidden layer whose node count
is determined via an evolutionary algorithm based on fitness proportionate selection.

Rivera et al. \cite{charte2019automating} also propose and evolutionary algorithm to
evolve autoencoders. Their work differs dramatically from both ours and that of
Suganuma et al. in that they use a uniform layer type throughout the network. They
also add a penalty to the fitness function that penalizes larger networks and thus
adding a selective pressure to simpler network architectures.

Sciuto et al. \cite{DBLP:journals/corr/abs-1902-08142} argue that a random search
policy outperforms many popular neural architecture search techniques. As our data
shows, this is not the case, at least for neural architecture search as applied to
finding autoencoder architectures for the tasks of manifold learning and image
denoising. In fact, we were somewhat surprised at how poorly random search performed
on the given tasks.

\section{Autoencoders}\label{sec:autoencoders}
\begin{figure}
\centering
  \includegraphics[width=.85\columnwidth]{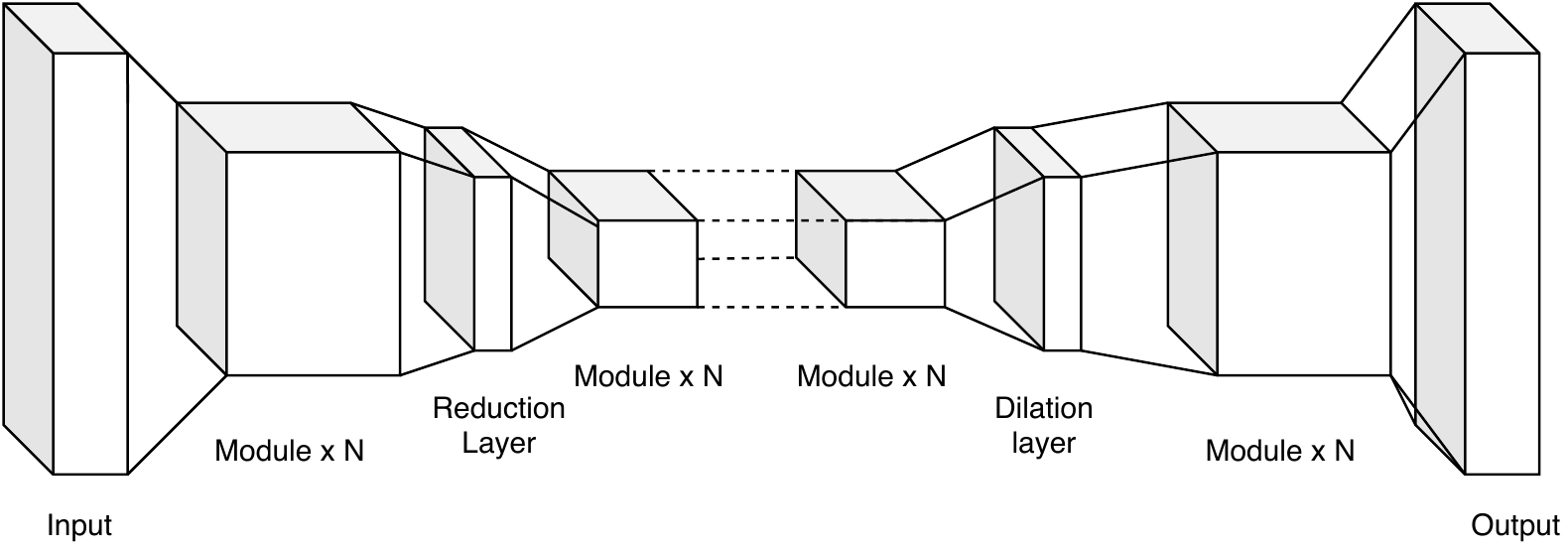}
  \caption{Diagram of modular autoencoder.}
  \label{fig:modular-ae}
\end{figure}

Autoencoders are a type of unsupervised learning technique consisting of two neural
networks, an \emph{encoder} network, denoted by $\psi_e$, and a \emph{decoder} network,
denoted by $\psi_d$, where the output of the encoder is fed into the
decoder. The two networks are trained together such that the output of the
decoder matches the input to the encoder. In other words, for an input
$\mathbf{x}$:
\[
  \mathbf{\hat{x}} = \psi_d(\psi_e(\mathbf{x}))
\]
and the goal is for the difference between $\mathbf{x}$ and
$\mathbf{\hat{x}}$ to be as small as possible, as defined by the loss function.
For a loss function $L$, the encoder and
decoder (collectively referred to as the autoencoder) optimize the problem given
by equation~\eqref{eq:ae-problem}

\begin{equation}\label{eq:ae-problem}
  \Psi(\mathbf{x}; \mathbf{\omega}_e, \mathbf{\omega}_d) =
\min_{\mathbf{\omega}_e, \mathbf{\omega}_d}L(\mathbf{x},
\psi_d(\psi_e(\mathbf{x}; \mathbf{\omega}_e); \mathbf{\omega}_d))
\end{equation}

where $\Psi(\mathbf{x}; \mathbf{\omega})$ represents the
encoder-decoder pair
\[
  \Psi(\mathbf{x}; \mathbf{\omega}) = (\psi_e (\mathbf{x}; \mathbf{\omega}_e), \psi_d(\mathbf{x}; \mathbf{\omega}_d))
\]
and $\mathbf{\omega}_e$ and
$\mathbf{\omega}_d$ are the weights of the encoder and decoder networks,
respectively. Typically the encoder network is a mapping to a lower-dimensional
space
\[
\psi_e : \mathbb{R}^d \rightarrow \mathbb{R}^{d'}
\]
and the decoder network is a mapping from the lower-dimensional space to the original
input space
\[
\psi_d : \mathbb{R}^{d'} \rightarrow \mathbb{R}^d
\]
where $d' \ll d$. The intuition behind this lower dimensional representation,
or embedding, is that the autoencoder transforms the input into a representation
that captures essential information and removes non-essential information. In
practice, the size of this lower-dimensional representation, which we refer to
as the intermediate representation, is a hyper-parameter of the autoencoder.

\begin{figure}
  \centering
  \includegraphics[width=\columnwidth]{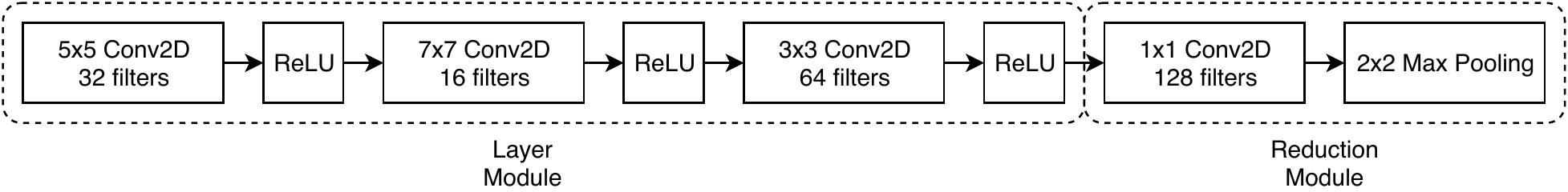}
  \caption{Example of a single layer module, followed by a reduction module.}
  \label{fig:layer-module}
\end{figure}

The autoencoders in this paper are constructed from repetitions of layer modules
separated by reduction (dilation) modules, as shown in
Figure~\ref{fig:modular-ae}.  A module is an ordered collection of layers,
separated by ReLU activations, as shown in
Figure~\ref{fig:layer-module}.
Layer
modules do not change the spatial dimensions of their input through the use of
zero padding. Reducing the spatial dimensions of the input is left to the
reduction module. Maintaining spatial dimension within layer modules allows us
to guarantee properly formed networks during the neural network construction
phase. A reduction module consists of a $1\times 1$ convolutional layer
followed by a $2\times 2$ max pooling layer with stride two. The $1\times 1$
convolutional layer doubles the depth of the input layer's filters, in an
attempt to reduce the information loss resulting from the strided max pooling
layer. The strided max pooling layer reduces each spatial dimension by a factor
of two, resulting in an overall reduction in information of 75\% (a quarter of
the original spatial information is retained). Doubling the number of filters
means the aggregate information loss is at least 50\% rather than 75\%. In the
decoder network, reduction modules are replaced with dilation modules. The
dilation modules replace the max pooling layer with a 2D convolutional transpose
layer \cite{10.1007/978-3-319-10590-1_53}.  The effect of this layer is to
expand the spatial dimension of its input.

Modular network designs, where the neural network is composed by repeating a
layer module a set number of times, are a common technique in neural architecture search
\cite{pmlr-v80-pham18a, 45826, DBLP:journals/corr/abs-1806-09055,
  DBLP:journals/corr/HaDL16, DBLP:journals/corr/abs-1708-05344,Koutnik:2013:ELN:2463372.2463509, DBLP:journals/corr/MiikkulainenLMR17,
DBLP:journals/corr/abs-1810-05749}. It can also be seen in many popular network architectures such as
Inception-v3 \cite{DBLP:journals/corr/SzegedyVISW15} and ResNets \cite{DBLP:conf/cvpr/HeZRS16}.
One of the primary advantages of using a
modular network architecture is that we are able to reduce the size of the
search space. For example, consider a modular network where the module has a
maximum size of five layers and is
repeated twice and there are two reduction layers. This network structure will
result in a maximum network of 30 layers; however, because the search is
performed at the \emph{module} level, we only search the space of five layer
networks. In our work we consider four different convolutional layers with four
different filter counts as well as a 2D dropout layer; this gives 17 different
options for each layer of the module. Using a modular network
architecture enables us to reduce the search space from $\sim 10^{85}$ to $\sim 10^{15}$.

\subsection{Image Denoising}
 Image denoising solves the problem
 \begin{equation}
   \Psi = \min_{\mathbf{\omega}_e, \mathbf{\omega}_d}L(\mathbf{X}, \psi_d(\psi_e(\mathbf{X} + \mathbf{\epsilon}; \mathbf{\omega}_e); \mathbf{\omega}_d))
\end{equation}
where the loss function, $L$, is typically the mean squared error (MSE)
loss shown in equation~\ref{eq:mse-ae}. We choose MSE in this setting because
it is a more appropriate choice for regression. If you consider the input and
output of the autoencoder is two separate functions, the goal of training is
to make them as similar as possible. While we could use binary cross entropy
from a technical point of view, we find it is better suited to classification
problems, of which autoencoders are decidedly not. On the other hand, one could
extend our algorithm to include the loss function as a parameter of the search,
as done by Rivera et al. \cite{charte2019automating}.
\begin{equation}\label{eq:mse-ae}
  L(\mathbf{X}, \psi_d(\psi_e(\mathbf{X} + \lambda\epsilon))) =
  \sum_{\mathbf{x}\in\mathbf{X}}||\mathbf{x} - \psi_d(\psi_e(\mathbf{x} + \epsilon))||^2
\end{equation}
and $\epsilon \sim N(0, \Sigma)$ where $\Sigma  = \sigma^2 \mathbb{I}$ and $\mathbb{I}$
is the identity matrix in $\mathbb{R}^{d \times d}$.
We tune the difficulty of the denoising
problem by modifying $\sigma$, larger $\sigma$ leads to noisier images. The intuition of denoising is
rather simple, learn
weights that allow the decoder to remove noise from a given image. There are a number
of applications of this technique, the most obvious application being image restoration.
However, initial work on denoising autoencoders actually used this technique as a
method of pre-training the neural networks and then fine-tuning the encoder to be used
in a classification task.

\subsection{Manifold Learning}
Manifold learning \cite{ma2011manifold, lin2008riemannian, talwalkar2008large}
is a form of non-linear dimensionality reduction.
Manifold learning algorithms assume the data, $\mathbf{X}$, is sampled from
a low-dimensional manifold $\mathcal{M} \subset \mathbb{R}^{d'}$
embedded in a higher dimensional space $\mathbb{R}^d$, where $d' \ll d$.
Some of the more popular
techniques are Isomap \cite{Tenenbaum2319}, locally-linear embeddings \cite{Roweis00nonlineardimensionality},
and Laplacian eigenmaps \cite{Belkin:2003:LED:795523.795528}.
Autoencoders are another technique for manifold learning.
In this case,  the encoder network $\psi_e$ is trained such that
for input $\mathbf{X} \subset \mathbb{R}^d$
\[
\psi_e : \mathbf{X} \subset \mathbb{R}^d \rightarrow \mathcal{M}
\]
and the decoder network is trained such that it is a mapping
\[
\psi_d : \mathcal{M} \rightarrow \mathbf{X} \subset \mathbb{R}^d
\]
This technique is useful in high-dimensional settings as a feature selection
technique, similar to Principal Component Analysis (PCA) \cite{Abdi:2010:PCA:3160436.3160440}
except that PCA is a linear dimensionality reduction technique and manifold
learning is non-linear. A major difference between linear and non-linear
manifold learning algorithms is that the linear algorithms attempt to preserve
global structure in the embedded data while the non-linear algorithms only
attempt to preserve local structure.

The primary challenge in manifold learning with autoencoders is two-fold:
we must choose reasonably performant neural network architectures for the
encoder and deocoder networks and we must also choose an appropriate dimension
$d'$. In this work we consider two $d'$ values of $d' = (1/4) d$ and
$d' = (1/8)d$, which we manually set for each experiment. The architecture of the
networks is found via evolutionary search. In practice, we would chose $d'$ as
the smallest $d'$ value that minimized the reconstruction loss
$L(\mathbf{X}, \psi_d(\psi_e(\mathbf{X})))$.

\section{Evolving Deep Autoencoders}\label{sec:nas}
The two main approaches to neural architecture search are based on reinforcement learning
or evolutionary algorithms. In this work we focus on the evolutionary approach, which
consists of two primary components: evolution and selection. The evolution step
is where new individuals (autoencoders in our case) are generated. The selection
step is where the algorithm determines which individuals should survive and
which individuals should be removed from the population. We use a generational
selection mechanism where a population of network architectures goes through
mutation, evaluation, and selection as a group. Specifically, we use $(\mu + \lambda)$
selection where a parent population of size $\mu$ generates $\lambda$ offspring
and we select the top $\mu$ of the $\mu + \lambda$ individuals.

\subsection{Network Construction}
As discussed in Section~\ref{sec:autoencoders}, the autoencoders we consider in
this paper consist of a layer module, followed by a
reduction layer made up of a $1\times 1$ convolution and a $2\times 2$ max
pooling layer---this structure may be optionally repeated, with each repetition
reducing the dimension of the intermediate representation by 50\%.
We represent the module as a list of sequentially connected layer objects,
encoded in a human-readable format. Each layer is defined by a layer token
and a filter count, separated by a colon. Layers are separated by commas. For
example, a module consisting of a $5\times 5$ convolution layer with 16 filters
and two $3 \times 3$ convolution layers with 32 filters would be represented:
\[
\texttt{5x5conv2d:16,3x3conv2d:32,3x3conv2d:32}
\]
This encoding is easy to work with in that it can be stored as a variable length
array of strings, which allows us to handle layer mutation via indexing and
addition of a layer by simply appending to the list.
Of course, this could be condensed to a numerical coding scheme to reduce the
size of the encoding. In our experience, using a more verbose encoding greatly
simplified debugging and analyzing the results.

Using a sequential network architecture simplifies the
construction process of the autoencoder when compared to a wide architecture
such as the Inception module \cite{DBLP:journals/corr/SzegedyVISW15}, where
the module's input can feed into multiple layers. Networks are constructed
by assembling $1\times 1$, $2\times 2$, $3\times 3$, $5\times 4$, and $7\times 7$ 2D
convolutional layers as well as a 2D dropout layer \cite{DBLP:conf/cvpr/TompsonGJLB15}
with dropout probability $p = 0.5$

We use convolutional layers, rather than dense layers, primarily because we use
color image inputs in our experiments, which have three color channels for each
pixel. Convolutional layers are useful for feature detection because they work
on patches of pixels in the input. Additionally, they are an effective technique
to reduce the total number of network parameters, resulting in smaller and
faster training networks.

Our algorithm can work with any layer or activation function types and it would
be appropriate to consider other layer types in different problem settings. For
example, in a feature selection setting with dense data vectors it would make
more sense to consider dense layers with evolving node counts rather than different
types of convolutional layers.

\begin{figure}
  \centering
  \includegraphics[width=0.5\columnwidth]{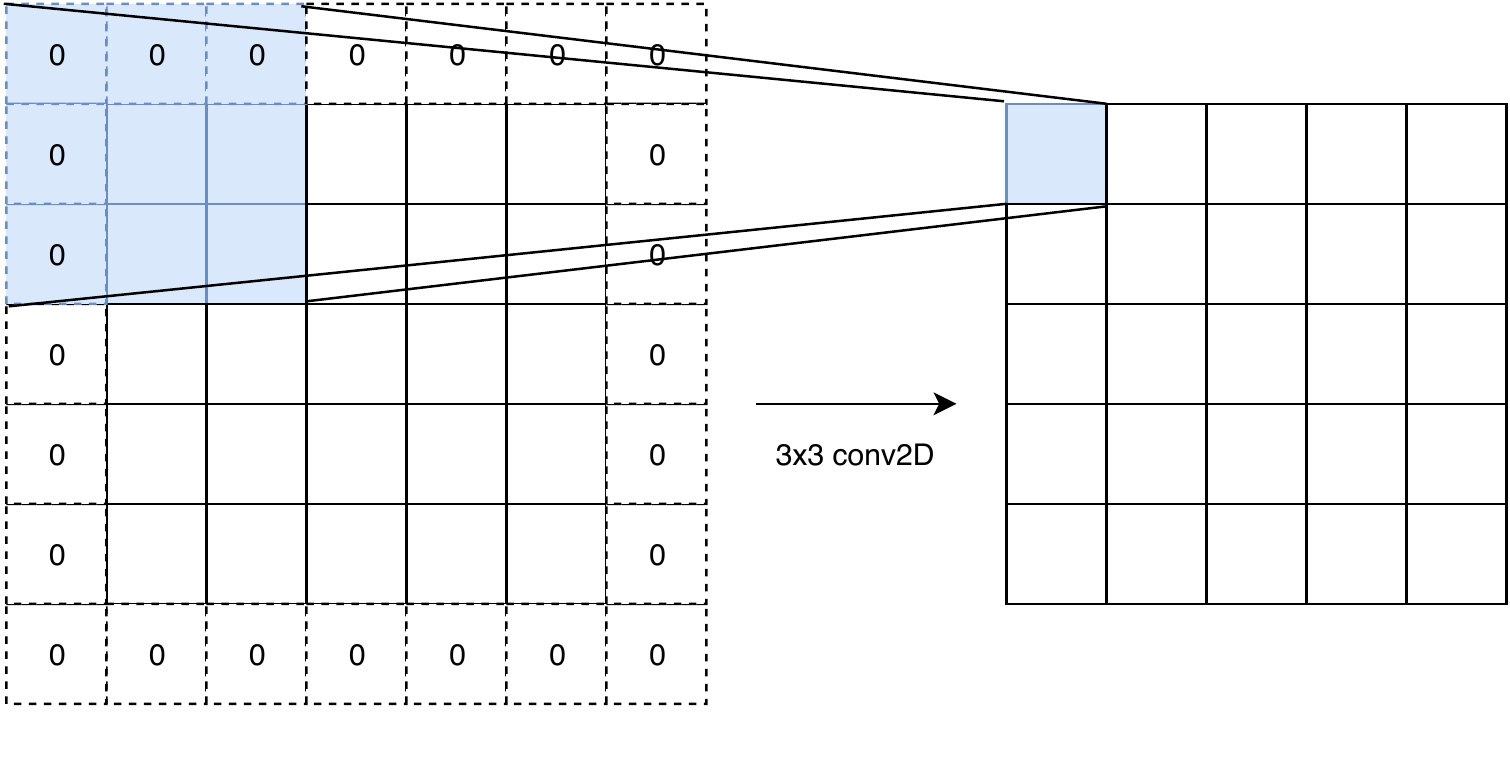}
  \caption{Illustration of a 2D $3\times 3$ convolutional layer with 0 padding to
    maintain the spatial dimensions of the input.}
  \label{fig:3x3conv2d}
\end{figure}
Figure~\ref{fig:3x3conv2d} illustrates a $3\times 3$ convolutional layer with zero padding.
We zero-pad all convolutional layers to maintain the spatial dimensions of the
input, using the reduction(dilation) layers to modify spatial dimensions.
Each layer can have 8, 16, 32, or 64 filters and the final output of the model is
passed through a $\tanh$ activation function.

\begin{algorithm}
  \caption{Evolutionary neural architecture search.}
  \label{alg:evo-nas}
  \begin{algorithmic}[1]
    \Procedure{EvolveNN}{$n, n_{\text{gen}}$}
    \State $P \gets$ new population of size $n$
    \For{ $i = 0$ to $n_{\text{gen}}$}
    \State \texttt{Mutate(}$P$\texttt{)}
    \For{ $p \in P$ }
    \If {$p$\texttt{.isEvaluated}}\label{alg:eval}
    \State \textbf{continue}
    \ElsIf { $p \in \texttt{cache.keys()}$}\label{alg:cache}
    \State $p.\texttt{setFitness(cache.get(}p\texttt{))}$
    \EndIf
    \State $t \gets \texttt{Task}(p)$
    \State Send task $t$ to broker
    \EndFor
    \State Wait for $|P|$ results from broker
    \State Update \texttt{cache} with received fitness values
    \State \texttt{sort(}$P$\texttt{)}
    \State $P \gets P[:n]$ \Comment{Select top $n$ individuals}
    \EndFor
    \EndProcedure
  \end{algorithmic}
\end{algorithm}

\begin{algorithm}
  \caption{Population evolution algorithm.}
  \label{alg:population-mutate}
  \begin{algorithmic}[1]
    \Procedure{Mutate}{$P$}
    \State $P' \gets P\texttt{.clone()}$
    \For{$p \in P$}
    \State $o \gets \texttt{RandomSample(}P \setminus p \texttt{)}$
    \State $u_1 \gets U(0,1)$
    \State $u_2 \gets U(0,1)$
    \If{$u_1 < .5$}
    \State $p \gets \texttt{MutateGenotype(}p\texttt{)}$ \Comment{Mutated clone of $p$}
    \EndIf
    \If{$u_2 < .5$}
    \State $o \gets \texttt{MutateGenotype(}o\texttt{)}$ \Comment{Mutated clone of $o$}
    \EndIf
    \State $c \gets \texttt{Crossover(}p, o\texttt{)}$
    \State \emph{// Only add $p$ and $o$ to $P'$ if they were mutated.}
    \State $P' \gets P' \cup \{p, o, c\}$
    \EndFor
    \State \textbf{return} $P'$
    \EndProcedure
  \end{algorithmic}
\end{algorithm}

\begin{algorithm}
  \caption{Mutation and crossover algorithms used by the genotypes.}
  \label{alg:mutate-crossover}
  \begin{algorithmic}[1]
    \Procedure{MutateGenotype}{$g$}
    \State $c \gets g\texttt{.clone()}$
    \State $z \gets \texttt{SampleUniform(}0, 1\texttt{)}$
    \If{$|c| < \texttt{MaxNumLayers} \textbf{and} z < 0.5$}
    \State $\texttt{AppendRandomLayer(}c\texttt{)}$
    \Else
    \State $\texttt{ReplaceRandomLayer(}c\texttt{)}$
    \EndIf
    \State \textbf{return} c
    \EndProcedure
    \Procedure{Crossover}{$g_1, g_2$}
    \State $i_1 \gets U(0, |g_1|$ \Comment{Random layer index for $g_1$}
    \State $i_2 \gets U(0, |g_2|$ \Comment{Random layer index for $g_2$}
    \State $c \gets g_1\texttt{.layers}[:i_1] + g_2\texttt{.layers}[i_2:]$
    \If{$|c| > \texttt{MaxNumLayers}$}
    \State \emph{//Drop layers exceeding size limit.}
    \State $c \gets c\texttt{.layers}[:\texttt{MaxNumLayers}]$
    \EndIf
    \State \textbf{return} $c$
    \EndProcedure
  \end{algorithmic}
\end{algorithm}

The sequential approach also has a major advantage in that it is guaranteed to
construct a valid network. Because each layer feeds into the next there will
always be a path from input to output. This is not the case when allowing
arbitrary connections between layers and letting evolution select these
layers.

\subsection{Network Evolution}
Network architectures are evolved, as described by Algorithm~\ref{alg:evo-nas},
by starting with a minimal layer module (e.g., a single convolutional layer) and
either mutating the module or by performing crossover with another layer
module. The specific algorithm used to evolve the population a single
generation is described in Algorithm~\ref{alg:population-mutate} while
Algorithm~\ref{alg:mutate-crossover} describes the specific algorithms
used to mutate individual genotypes and perform crossover between two
genotypes.
Mutation works by either appending a layer or modifying an existing
layer. Crossover between two modules involves taking two network architectures
and splicing them together. All of this is performed by the model, which is
described in greater detail in Section~\ref{sec:sys-arch}.  The autoencoder is
packaged into a network task Protocol Buffer and sent to the Broker, which
forwards them to a worker. The workers perform the task of training and
evaluating the autoencoder, where evaluation is performed on validation data
that the network as not previously seen. The validation loss is packaged in a
result Protocol Buffer and sent back to the Broker, who then forwards the result
to the model. We define fitness as the reciprocal of the validation loss, which
has the nice property that low loss results in high fitness.

One notable aspect of our evolution algorithm is that we make it very likely
one offspring is produced from each genotype in the parent
population. We force crossover as long as the two selected
genotypes have more than one layer each. This is similar to the forced mutation
of Suganuma et al. \cite{pmlr-v80-suganuma18a}. The advantage is that we are
constantly exploring new architectures; however, this approach also results in
very large offspring populations that can be as large as $3\times$ the size of
the parent population. If the parents were mutated during crossover their
mutated selves are added into the original population. This differs from a more
classical evolutionary algorithm in that mutation and crossover are strongly
encouraged and forced. This adds variance to the overall population fitness,
which we counter by using $(\mu + \lambda)$ selection, making sure we always
maintain the best performing network architectures.

To cope with potentially large offspring population sizes, we make two
modifications to the standard evolution algorithm to improve efficiency by
reducing the total amount of work. These can be seen in lines~\ref{alg:eval}
and~\ref{alg:cache} in Algorithm~\ref{alg:evo-nas}. In line~\ref{alg:eval}, the
model checks if the candidate autoencoder architecture (referred to as a
genotype) has previously been evaluated. If the genotype was previously
evaluated the model decides not to send it off for re-evaluation. This is possible because
we make genotypes immutable---if a genotype is selected for mutation or
crossover, a copy is made and mutated rather than modifying the original
genotype.
This modification alone can save anywhere from 25\% (offspring
population size $3\times$ that of the parent population) to 50\% (offspring
population size $2\times$ that of the parent population) because we use $(\mu +
\lambda)$ selection, so the parent population is always included when evaluating
the individuals of a population.

Similarly, in line~\ref{alg:cache}, we check if the
architecture of the genotype as been seen previously.
Caching previously seen architectures and their respective fitness is
particularly important because of the stochastic nature of the evolutionary
search algorithm. We noticed during the experiments, especially at later
generations of evolution when the population has started to homogenize, that
previously encountered architectures would be rediscovered. This is expected
because evolutionary search stochastically explores the neighborhood of
the current position, which means it may explore previously seen locations
(architectures) simply due to chance.

Let $\psi(\mathbf{x}; \mathbf{\omega})$ represent a trained neural network with a fixed architecture.
We denote training data as $\textbf{X}_{\text{tr}}$ and validation data as $\textbf{X}_\text{val}$.
We define an autoencoder $\Psi = (\psi_e, \psi_d)$ as a tuple containing
an encoder network, $\psi_e$, and a decoder network, $\psi_d$.
Formally, the problem we solve in this work is shown in equation~\eqref{eq:problem},
\begin{equation}\label{eq:problem}
  \Psi^* = \arg\min_{\psi \in \mathcal{A}} L(\textbf{X}_{\text{val}},
  \psi_d(\psi_e(X_{\text{val}})))
\end{equation}
The optimal autoencoder for the given problem is given by $\Psi^*$,
where optimality is defined with respect to an architecture in the space of
all neural network architectures, $\mathcal{A}$. The loss function $L$
is mean-squared error, as defined in equation~\eqref{eq:mse}, where
$\mathbf{X}, \mathbf{Y} \in \mathbb{R}^{n\times d}$.
\begin{equation}\label{eq:mse}
  L(\mathbf{X}, \mathbf{Y}) = \frac{1}{n}\sum_{i=1}^n||\mathbf{x} - \mathbf{y}||^2
\end{equation}

%

\begin{figure}
  \centering
  \includegraphics[width=.5\columnwidth]{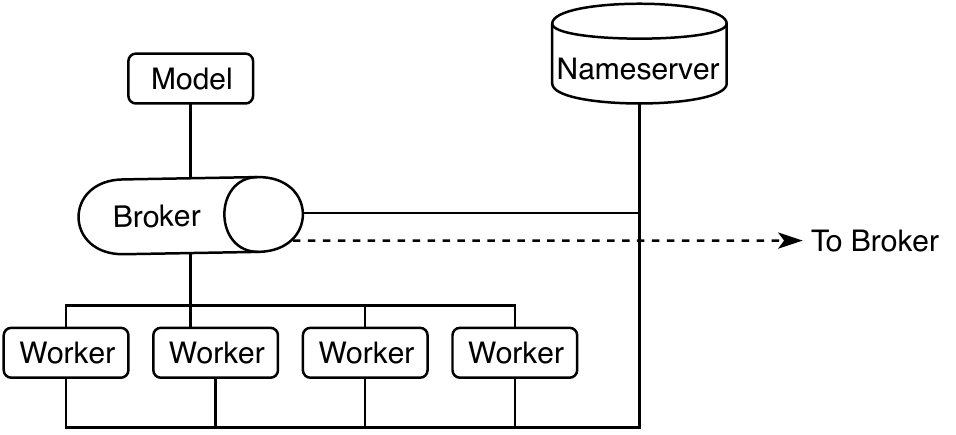}
  \caption{Overview of the system architecture.}
  \label{fig:sys-arch}
\end{figure}

\subsection{System Architecture}\label{sec:sys-arch}
We designed a system to efficiently find and evaluate deep autoencoders using
the evolutionary algorithm described in Algorithm~\ref{alg:evo-nas}. We use
the system proposed in \cite{hajewski-bdl} and shown in Figure~\ref{fig:sys-arch},
which consists of a model, one or more
brokers, an arbitrary number of workers, and a nameserver. Flow of information
through the system is rather simple -- data flows from the model to the workers
and then back to the model, all through the broker. Communication within the
system is handled via remote procedure calls (RPC) using gRPC \cite{grpc} and messages are serialized
using Protocol Buffers \cite{Varda2008}. The system infrastructure is
written in Go while the model and worker implementations are written in
Python and are availabe on GitHub\footnote{
\url{https://github.com/j-haj/brokered-deep-learning}}.
This is possible because of gRPC---we specify the system APIs in
gRPC's interface description language (IDL) and then implement them in Go
and Python. The model and workers make use of the Python stubs and clients
generated by gRPC from the API specification. The system moves the serialized
Python objects from the model to the worker without having to worry about
the contents of the messages. The advantage is that we can build the system
infrastructure in a statically type-checked language such as Go while we
can use Python and PyTorch \cite{paszke2017automatic} to build, train, and evaluate the autoencoder
architectures. This is particularly important for the broker implementations;
using Go gives the broker a high level of concurrency and performance when
compared to Python.

Heartbeat messages are RPCs used by a client to inform a server that the
client is still functioning. This is important when a client requests a
resource within the system, such as a worker requesting a task from a
broker. The resource may be lost if the client fails and no heartbeat
mechanism is used. The heartbeat message informs the server if the
client has failed, allowing the server to properly handle the shared
resources.

\paragraph{Model}
The model drives the evolutionary search for autoencoders by running the
evolutionary algorithm that designs the layer modules. Once a population of
layer modules has been generated (e.g., mutated or mated via crossover),
the model packages the individual designs into tasks and sends them to the
broker via RPC\@. The model then waits for the broker to send back the results
from each of the layer module designs.

\paragraph{Broker}
The broker forms the data pipeline of the system by moving data between the
model and the workers. Unevaluated network architectures received from the
model are stored in a
queue. These network
architectures--referred to as tasks--are removed from the queue and sent
to a worker when a task request RPC is received from the respective worker.
At this point the task is moved to a pending-completion state until the
finished result is returned from the worker, at which point the result is
forwarded to the model.


Brokers can connect (or link) with other brokers to share available compute resources.
When connected with another broker, the broker stores shared tasks in a work
queue, giving preference to its own tasks (referred to as \emph{owned tasks}).
Linked brokers are typically in different data centers because a single broker is
capable of handling a large number of connected workers. Each network architecture
evaluation takes on the order of minutes, giving the broker ample time to distribute
tasks to other workers.

\paragraph{Worker}
The worker is the computational engine of the system. Workers register with a broker
when they request a task from the broker but are otherwise able to join or leave
the system at-will. A worker begins a heartbeat RPC with the broker from which it
requested a task. This heartbeat RPC ends once the task is complete and returned
from the broker. In this sense, workers lease tasks from the broker. As a result,
if a worker leaves the system after returning a task, no other components in the
system will know the worker left and there will be no state within the system
waiting for additional information from the worker.

\paragraph{Nameserver}
The nameserver is a central store for the addresses of the brokers in the system.
Models or workers can query the central store for a broker address prior to joining
the system. This is not strictly required for the system, but improves the
scalability of the system as more nodes are added. Rather than manually specifying
the broker address for each worker, which may vary if there are many brokers, each
worker can query the nameserver and get the address of a broker to connect to.
Similarly, brokers can query the nameserver for the addresses of other brokers with
whom they can form a link for work-sharing. As a result, the nameserver establishes
a heartbeat with connected brokers. Brokers are removed from the nameserver's list
of available brokers when they fail to send a heartbeat within a specified time
window. If the heartbeat is late, the nameserver will reply with a \texttt{reconnect}
request to the broker, causing the broker to re-register with the nameserver.


\begin{table*}
  \centering
  \caption{Fitness of top 3 found architectures for image denoising, trained for 20 epochs.}\label{tab:top-3}
\begin{tabular}{cccc}
\toprule
\textbf{\# of Epochs During Search} & \textbf{Rank} & \textbf{Architecture} & \textbf{Fitness} \\ \midrule
\multirow{3}{*}{2} & 1 & 64-3x3conv2d  & 6.10 \\ \cline{2-4}
                   & 2 & 64-7x7conv2d  & 5.97 \\ \cline{2-4}
                   & 3 & 64-7x7conv2d - 64-3x3conv2d & 5.86 \\ \hline
\multirow{3}{*}{5} & 1 & 64-5x5conv2d  & 6.02  \\ \cline{2-4}
                   & 2 & 64-7x7conv2d  & 5.97  \\ \cline{2-4}
                   & 3 & 64-7x7conv2d - 64-5x5conv2d  & 5.96  \\ \bottomrule
Random & - & - & $1.8 \pm 1.52$\\\bottomrule
\end{tabular}
\end{table*}

\section{Experiments}\label{sec:experiments}
We explore two areas of application in the experiments: manifold learning and image denoising.
For all experiments we use the STL10 \cite{DBLP:journals/jmlr/CoatesNL11} dataset. We
chose this dataset for its difficulty---the images are higher resolution than the
CIFAR-10/100 \cite{cifar10} datasets, allowing more reduction modules. Additionally,
the higher-resolution images are better suited for deeper, more complex autoencoder
architectures.

The layer module is restricted to at most ten layers. This gives a sufficiently large
search space of around 10 billion architectures. We also compare the found architectures
against random search. Over-fitting is a challenge with deep autoencoders, so we restrict
our models to only one layer module, rather than repeating the layer module multiple times.
In our initial experiments we found autoencoders with multiple layer modules performed
dramatically worse (due to over-fitting) than their single layer module counterparts.

The networks are implemented and trained using the PyTorch \cite{paszke2017automatic}
framework. Evolutionary search is performed over 20 generations with a population
size of 10 individuals. We use $(\mu + \lambda)$ selection as described in
Algorithm~\ref{alg:evo-nas}.

\paragraph{Random Search}
The random search comparisons are performed by sampling from $U[1,10]$,
and using the sampled integer as the number of layers in the layer module. Each
layer is determined by sampling a random layer type and a random number of filters.
The number of layer modules and the number of reductions are hyper-parameters and set
to the same values as the comparison architectures found via evolutionary search.
We sample and evaluate 30 random architectures for each experiment.

\paragraph{Image Denoising}
We used the image denoising experiments to test how well the evolutionary search
algorithm performs when the candidate network architectures are trained for either
two or five epochs. If the search can get away with fewer training epochs it will
save resources, allowing the algorithm to explore more architectures. We set
$\sigma^2 = 1/3$ in these experiments because we found that too much noise would
cause the system to collapse and output blank images.

\paragraph{Manifold Learning}
In the manifold learning setting we focus on a restricted set of reduced dimensions,
namely those dimensions that are reduced by a factor of $(1/2)^k$ for $k = 2$ and $3$.
This greatly simplifies
the construction of both the encoder and decoder networks. Each reduction module,
consisting of a $1\times 1$ 2D convolutional layer followed by a stride 2, $2\times 2$
max pooling layer, reduces the each of the spatial dimensions by a factor of two.
Thus the input dimension is reduced by 75\% and 87.5\%, respectively.
This is a fairly drastic reduction in dimension and limits the total number of
reduction modules used in a given autoencoder architecture.
\begin{figure*}
  \centering
  \begin{subfigure}[t]{0.45\textwidth}
    \includegraphics[width=1\linewidth]{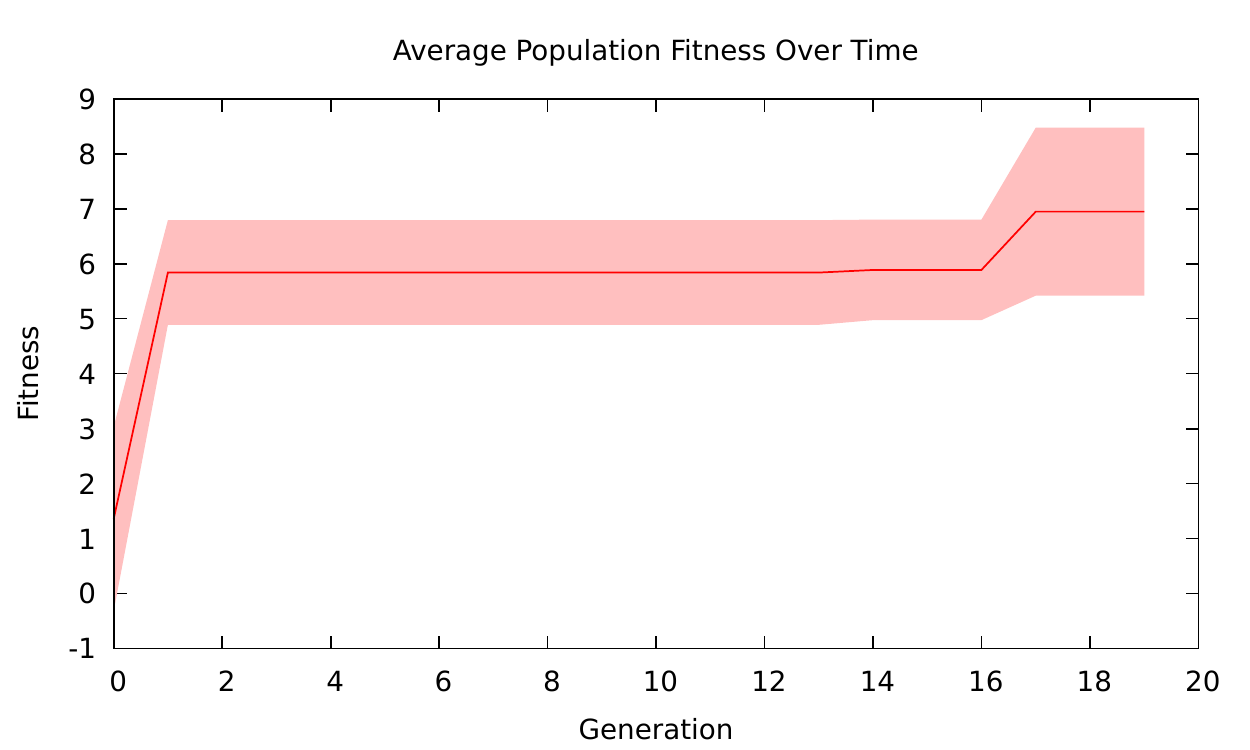}
    \caption{Average fitness over 20 generations using two epochs during training.}
    \label{fig:fuzz2}
  \end{subfigure}
  \begin{subfigure}[t]{0.45\textwidth}
    \includegraphics[width=1\linewidth]{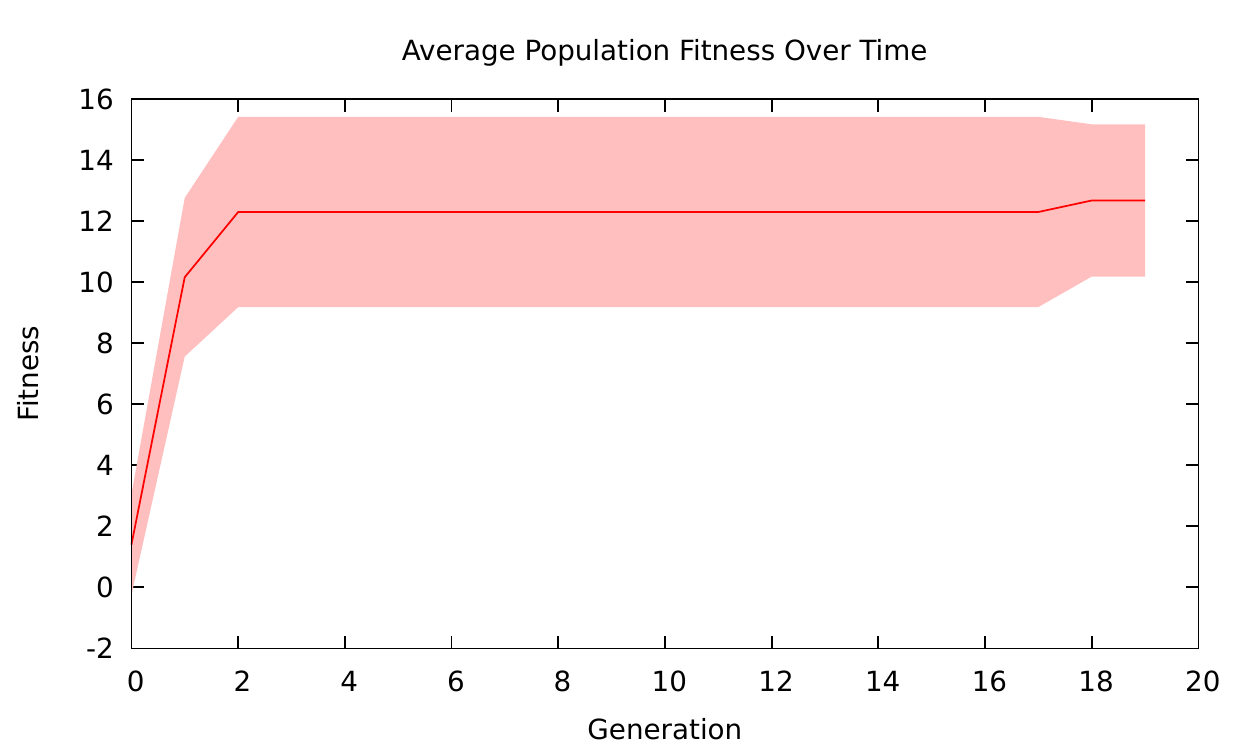}
    \caption{Average fitness over 20 generations using five epochs during training.}
    \label{fig:fuzz5}
  \end{subfigure}
  \caption{Average fitness over 20 generations.}
  \label{fig:fuzz-avg}
\end{figure*}
\section{Discussion}\label{sec:results}
We present the results form our experiments along with a discussion of their
implications. All experiments were performed on Amazon Web Services (AWS)
using Nvidia K80 GPUs. Fitness is defined as the reciprocal of the loss, lower
loss leads to greater fitness. We train the final architectures for 20 epochs
in the image denoising experiments and 40 epochs in the manifold learning
experiments.
In practice, this is a very small number of epochs for
a production model; however, we kept this number low as it did not affect the
results and reduced costs.

\paragraph{Random Search}
Random search performed dramatically worse, on average, than evolutionary search
on both the manifold learning and denoising tasks. This is unsurprising given the
size of the search space, but differs from the claims of \cite{DBLP:journals/corr/abs-1902-08142}.
We train each random architecture for 20 epochs in the image denoising experiments
and 40 epochs in the manifold learning experiments.
\paragraph{Image Denoising}

Table~\ref{tab:top-3} summarizes the results from the denoising experiments.
We list the top three network architectures for the 2-epoch and 5-epoch searches, along
with their fitness values after 20 epochs of training. Interestingly, both the
2-epoch and 5-epoch approaches found similar architectures and, perhaps more surprisingly,
their third place architectures were combinations of the second and
first place architectures.

It is unsurprising that both searches favored smaller architectures---the smaller
epoch count means networks that train faster will have a better evaluation on the
validation data. Smaller networks typically train faster than larger networks with
respect to number of epochs, so it makes sense that these networks would rank
higher during our searches using a limited number of epochs.
On the other hand,
over-fitting is an issue with the larger networks so favoring smaller layer module
architectures had a positive impact on the overall performance of the autoencoder.

Figure~\ref{fig:fuzz-avg} shows the average fitness across 20 generations for both
the 2-epoch and 5-epoch searches. Both graphs share two similar features: they
plateau at the second generation and maintain that plateau until around the 18th
generation, when they both find better architectures. Although the 5-epoch graph
has an unsurprisingly higher fitness, it is surprising that both approaches appear
to improve at about the same pace. This reinforces the idea that the 2-epoch search
does a decent job at exploring the space.

\begin{figure*}
  \centering
  \begin{subfigure}[b]{0.75\textwidth}
    \includegraphics[width=1\linewidth]{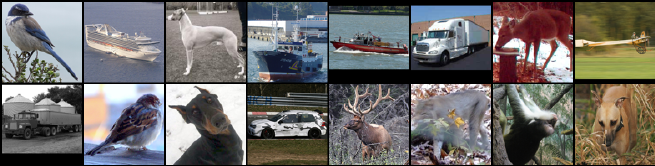}
    \caption{Original images (STL10 dataset).}
    \label{fig:fuzz-unfuzzed}
  \end{subfigure}
  \begin{subfigure}[b]{0.75\textwidth}
    \includegraphics[width=1\linewidth]{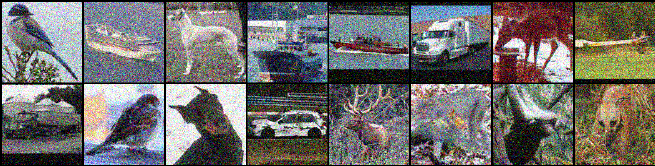}
    \caption{Noised input images.}
    \label{fig:fuzz-fuzzed}
  \end{subfigure}
  \begin{subfigure}[b]{0.75\textwidth}
    \includegraphics[width=1\linewidth]{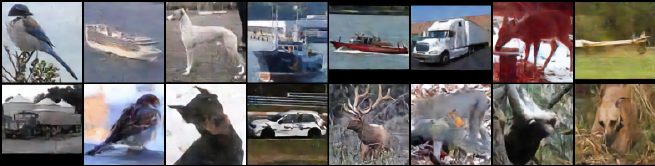}
    \caption{Denoised output images (20 epochs of training).}
    \label{fig:fuzz-denoised}
  \end{subfigure}
  \caption{Denoised images on unseen data.}
  \label{fig:denoise-best}
\end{figure*}

The un-noised input image is shown in Figure~\ref{fig:fuzz-unfuzzed},
the noised input images shown in Figure~\ref{fig:fuzz-fuzzed}, and the denoised
images shown in Figure~\ref{fig:fuzz-denoised}.

%
%

\paragraph{Manifold Learning}
\begin{table*}[t]
  \centering
  \caption{Fitness of top 3 found architectures for manifold learning, trained for 40 epochs.}\label{tab:man-top-3}
\begin{tabular}{cccc}
\toprule
\textbf{\# of Reduction Modules} & \textbf{Rank} & \textbf{Architecture} & \textbf{Fitness} \\ \midrule
\multirow{4}{*}{2}   & 1  & 64-5x5conv2d   & 9.05   \\ \cline{2-4}
                     & 2  & 32-3x3conv2d   & 6.75   \\ \cline{2-4}
                     & 3  & 32-7x7conv2d  &  4.99 \\ \cline{2-4}
                     & - & Random & $1.17 \pm 1.22$ \\\hline
\multirow{4}{*}{3}   & 1  & 64-3x3conv2d   & 3.42  \\ \cline{2-4}
                     & 2  & 64-5x5conv2d   & 3.38   \\ \cline{2-4}
                     & 3  & 32-3x3conv2d - 64-5x5conv2d   & 2.4 \\ \cline{2-4}
                     & - & Random & $0.36 \pm 0.30$ \\\bottomrule
\end{tabular}
\end{table*}

\begin{figure}[t]
  \centering
  \includegraphics[width=0.7\columnwidth]{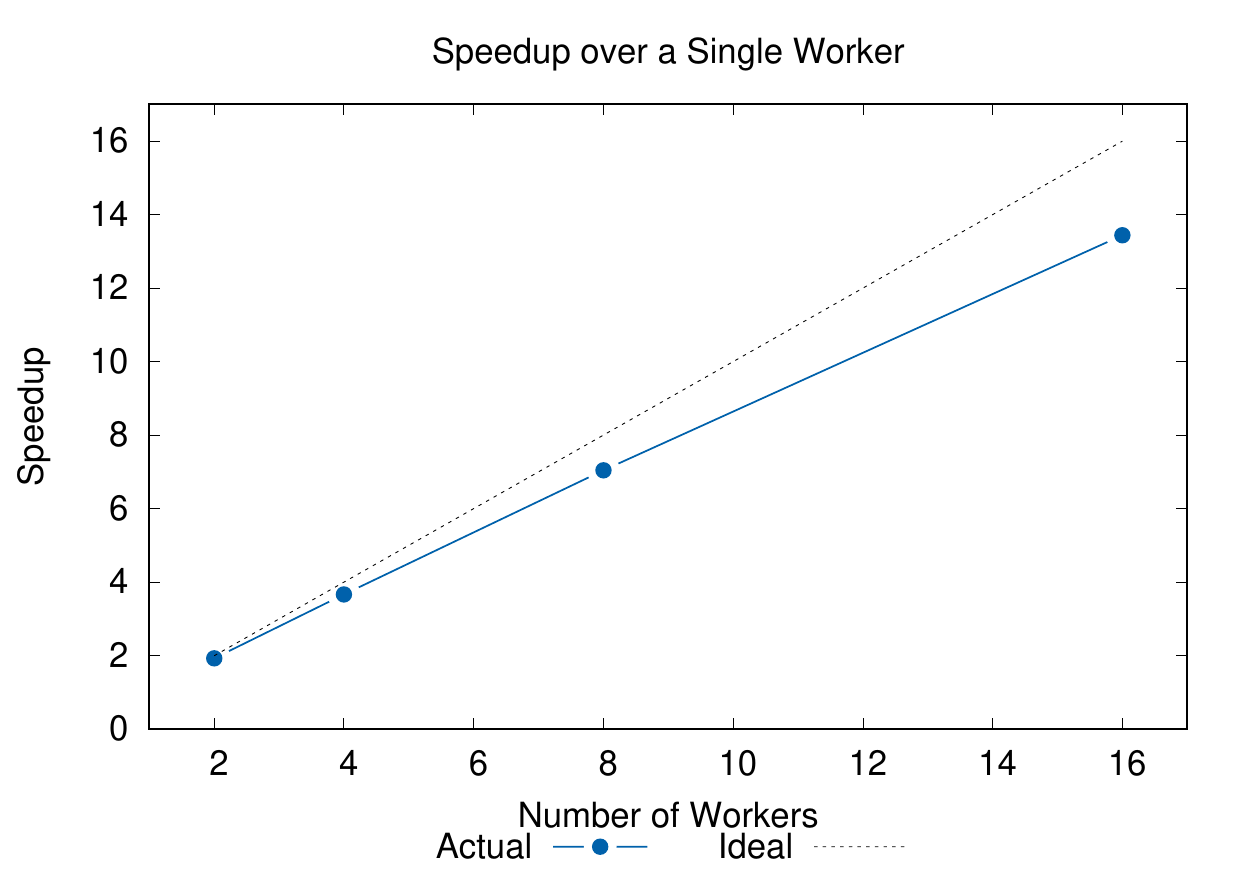}
  \caption{Scaling factors with respect to a single worker. These
    are calculated as the geometric average of the number of candidate
    architectures evaluated per generation, across five generations.}
  \label{fig:scaling}
\end{figure}

\begin{table*}
  \centering
  \caption{Autoencoder architecture for Dropout example.}
  \label{tab:ae-dropout}
  \begin{tabular}{lc}\toprule
    \textbf{Image Effect} & \textbf{Layer Module Architecture}\\\midrule
    Greyscale & 16-3x3conv2d -- Dropout2D -- 32-1x1conv2d -- 2x2max-pool\\
    Color & 64-7x7conv2d -- Dropout2D -- 32-7x7conv2d -- 2x2max-pool\\\bottomrule
  \end{tabular}
\end{table*}

Table~\ref{tab:man-top-3} summarizes the results from performing evolutionary search
on autoencoders using two and three reduction modules compared to random search.
The search was performed over 20 generations, as is done in the denoising experiments.
Because these networks are larger, we train them for 40 epochs instead of the 20
epochs used in the denoising experiments. Similar to the  denoising experiments,
random search performs worse than evolutionary search in all scenarios.

A single $5\times 5$ 2D convolutional layer as the layer module ranked highly in
both the two and three reduction experiments at first and second place, respectively.
As shown in Table~\ref{tab:man-top-3}, the 5x5conv2d layer with 64 filters performed
nearly 50\% better than the runner-up
3x3conv2d layer in the two reduction experiment. Interestingly, the 3x3conv2d and
5x5conv2d layers performed nearly identically in the three reduction experiment
with the 3x3conv2d layer taking a slight lead. Perhaps more notably, the 5x5conv2d
architecture performed nearly an order of magnitude better than random search in
both experiments.

\begin{figure*}
  \centering
  \begin{subfigure}[b]{0.75\textwidth}
    \includegraphics[width=1\linewidth]{img/en_img_10}
    \caption{Original input images (STL10 dataset).}
  \end{subfigure}
  \begin{subfigure}[b]{0.75\textwidth}
    \includegraphics[width=1\linewidth]{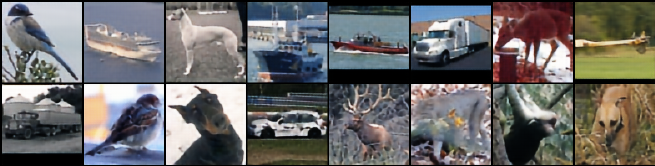}
    \caption{Reconstructed image from autoencoder using two reduction layers.
      The original representation is reduced by 75\%.}
    \label{fig:manifold-2red}
  \end{subfigure}
  \begin{subfigure}[b]{0.75\textwidth}
    \includegraphics[width=1\linewidth]{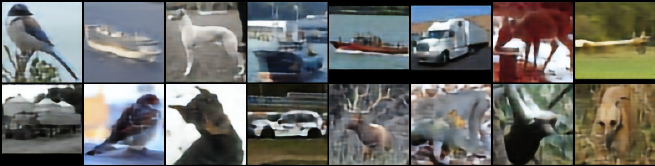}
    \caption{Reconstructed image from autoencoder using three reduction layers.
      The original representation is reduced by 87.5\%.}
    \label{fig:manifold-3red}
  \end{subfigure}
  \caption{Comparison of best architectures found after 20 generations.}
  \label{fig:manifold-res}
\end{figure*}

See the appendix for a demonstration of the results of the best autoencoder architectures
found after 20 generations of search. Here we compare the resulting images of two and three
reduction autoencoders.
The reconstructed images using the two
reduction autoencoder is shown in Figure~\ref{fig:manifold-2red} while the
reconstructed images using the three reduction autoencoder is shown in Figure~\ref{fig:manifold-3red}.
\paragraph{System Scalability}
Figure~\ref{fig:scaling} shows the scalability of the system. We were only able
to test the system with a maximum of 16 nodes due to resource constraints.
The system shows near linear scaling---the curve falls off from linear not due
to scalability issues with the system architecture but due to node failures.
We would occasionally experience GPU out-of-memory errors and this was more
prevalent with larger node counts. Using an orchestration system such as
Kubernetes \cite{43826} or Apache Mesos \cite{Hindman:2011:MPF:1972457.1972488}
would improve these scaling numbers as well as making the system more resilient.
\paragraph{Impact of Dropout Layers}
Dropout layers have an interesting, if not somewhat unsurprising, effect
on the output of the autoencoder. We give an example in the appendix
of two different networks, whose architectures are described in
Table~\ref{tab:ae-dropout}, after 10 epochs of training. The output of the
first network is greyscale
while the output the second network
is color (although it does exhibit some greyscale properties on some images).
We use a 2D dropout layer \cite{DBLP:conf/cvpr/TompsonGJLB15} to regularize
the autoencoders---during our experiments we noticed it was fairly easy to
overfit the data so we decided to add a dropout layer to the list of potential
layers used by the evolution algorithm. Note that, as described in
\cite{DBLP:conf/cvpr/TompsonGJLB15}, the 2D dropout layer drops out an entire
channel at random during training. Despite both networks having a 2D dropout
layer in the same position, the network that produces color output
has a greater learning capacity due to its larger parameter count. In other
words, this network
has more filters (and thus more parameters) and uses these additional filters
to store additional learned features of the dataset. This is what appears to
the characteristic that allows the
network to maintain coloring in the images.
Somewhat surprisingly given the ease with which we noticed the autoencoders would
overfit the data, none of the best architectures utilized a dropout layer.

\begin{figure*}
  \centering
  \begin{subfigure}[b]{0.75\textwidth}
    \includegraphics[width=1\linewidth]{img/en_img_10}
    \caption{Input data (STL10 dataset)}
    \label{fig:input-data-d1}
  \end{subfigure}

  \begin{subfigure}[b]{0.75\textwidth}
    \includegraphics[width=1\linewidth]{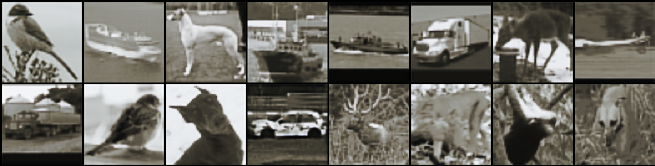}
    \caption{Output from decoder after 10 epochs and three layer, layer module of
      16-3x3conv2d -- Dropout2D -- 32-1x1conv2d, as described in Table~\ref{tab:ae-dropout}.}
    \label{fig:output-data-d1}
  \end{subfigure}

  \begin{subfigure}[b]{0.75\textwidth}
    \includegraphics[width=1\linewidth]{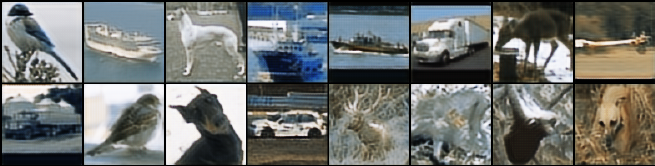}
    \caption{Output from decoder after 10 epochs and three layer, layer module of
      64-7x7conv2d -- Dropout2D -- 32-7x7conv2d, as described in Table~\ref{tab:ae-dropout}.}
    \label{fig:output-data-d2}
  \end{subfigure}

  \caption{Three layer module with a 2D dropout layer in the middle. Color
  information is preserved after 10 epochs of training.}
  \label{fig:dropout}
\end{figure*}

\section{Conclusion}
We have introduced an efficient and scalable system for neural architecture
search and demonstrated its effectiveness in designing deep autoencoders
for image denoising and manifold learning. An interesting avenue for future
work would be to extend the evolutionary algorithm based on the observations
of the denoising experiments---rather than search a fixed space for all
generations, use the first few generations to find smaller architectures that
work well. After these architectures are found, reduce the search space to
combinations of these architectures. This is similar, in a sense, to the work
of \cite{DBLP:journals/corr/abs-1711-00436}. Other important future work is
improving individual worker efficiency, allowing the system to achieve similar
performance using fewer workers. Although the large number of workers enables
fast search of network architectures, each worker consumes valuable resource
(e.g., monetary, compute, environmental, etc.). This work is important
in enabling self-improving and automatic machine learning.

\bibliographystyle{unsrt}  
\bibliography{nas_ae}

\begin{thebibliography}{10}

\bibitem{45826}
Barret Zoph and Quoc~V. Le.
\newblock Neural architecture search with reinforcement learning.
\newblock 2017.

\bibitem{DBLP:conf/icml/RealMSSSTLK17}
Esteban Real, Sherry Moore, Andrew Selle, Saurabh Saxena, Yutaka~Leon Suematsu,
  Jie Tan, Quoc~V. Le, and Alexey Kurakin.
\newblock Large-scale evolution of image classifiers.
\newblock In Doina Precup and Yee~Whye Teh, editors, {\em Proceedings of the
  34th International Conference on Machine Learning, {ICML} 2017, Sydney, NSW,
  Australia, 6-11 August 2017}, volume~70 of {\em Proceedings of Machine
  Learning Research}, pages 2902--2911. {PMLR}, 2017.

\bibitem{DBLP:journals/corr/abs-1711-00436}
Hanxiao Liu, Karen Simonyan, Oriol Vinyals, Chrisantha Fernando, and Koray
  Kavukcuoglu.
\newblock Hierarchical representations for efficient architecture search.
\newblock {\em CoRR}, abs/1711.00436, 2017.

\bibitem{ma2011manifold}
Yunqian Ma and Yun Fu.
\newblock {\em Manifold learning theory and applications}.
\newblock CRC press, 2011.

\bibitem{talwalkar2008large}
Ameet Talwalkar, Sanjiv Kumar, and Henry Rowley.
\newblock Large-scale manifold learning.
\newblock In {\em 2008 IEEE Conference on Computer Vision and Pattern
  Recognition}, pages 1--8. IEEE, 2008.

\bibitem{Vincent:2008:ECR:1390156.1390294}
Pascal Vincent, Hugo Larochelle, Yoshua Bengio, and Pierre-Antoine Manzagol.
\newblock Extracting and composing robust features with denoising autoencoders.
\newblock In {\em Proceedings of the 25th International Conference on Machine
  Learning}, ICML '08, pages 1096--1103, New York, NY, USA, 2008. ACM.

\bibitem{DBLP:journals/jmlr/VincentLLBM10}
Pascal Vincent, Hugo Larochelle, Isabelle Lajoie, Yoshua Bengio, and
  Pierre{-}Antoine Manzagol.
\newblock Stacked denoising autoencoders: Learning useful representations in a
  deep network with a local denoising criterion.
\newblock {\em J. Mach. Learn. Res.}, 11:3371--3408, 2010.

\bibitem{pmlr-v80-pham18a}
Hieu Pham, Melody Guan, Barret Zoph, Quoc Le, and Jeff Dean.
\newblock Efficient neural architecture search via parameters sharing.
\newblock In Jennifer Dy and Andreas Krause, editors, {\em Proceedings of the
  35th International Conference on Machine Learning}, volume~80 of {\em
  Proceedings of Machine Learning Research}, pages 4095--4104,
  Stockholmsmässan, Stockholm Sweden, 7 2018. PMLR.

\bibitem{DBLP:journals/corr/abs-1806-09055}
Hanxiao Liu, Karen Simonyan, and Yiming Yang.
\newblock {DARTS:} differentiable architecture search.
\newblock {\em CoRR}, abs/1806.09055, 2018.

\bibitem{DBLP:journals/corr/HaDL16}
David Ha, Andrew~M. Dai, and Quoc~V. Le.
\newblock Hypernetworks.
\newblock {\em CoRR}, abs/1609.09106, 2016.

\bibitem{DBLP:journals/corr/abs-1708-05344}
Andrew Brock, Theodore Lim, James~M. Ritchie, and Nick Weston.
\newblock {SMASH:} one-shot model architecture search through hypernetworks.
\newblock {\em CoRR}, abs/1708.05344, 2017.

\bibitem{Kyriakides:2018:NAS:3200947.3208068}
George Kyriakides and Konstantinos~G. Margaritis.
\newblock Neural architecture search with synchronous advantage actor-critic
  methods and partial training.
\newblock In {\em Proceedings of the 10th Hellenic Conference on Artificial
  Intelligence}, SETN '18, pages 34:1--34:7, New York, NY, USA, 2018. ACM.

\bibitem{Koutnik:2013:ELN:2463372.2463509}
Jan Koutn\'{i}k, Giuseppe Cuccu, J\"{u}rgen Schmidhuber, and Faustino Gomez.
\newblock Evolving large-scale neural networks for vision-based reinforcement
  learning.
\newblock In {\em Proceedings of the 15th Annual Conference on Genetic and
  Evolutionary Computation}, GECCO '13, pages 1061--1068, New York, NY, USA,
  2013. ACM.

\bibitem{DBLP:journals/corr/MiikkulainenLMR17}
Risto Miikkulainen, Jason~Zhi Liang, Elliot Meyerson, Aditya Rawal, Daniel
  Fink, Olivier Francon, Bala Raju, Hormoz Shahrzad, Arshak Navruzyan, Nigel
  Duffy, and Babak Hodjat.
\newblock Evolving deep neural networks.
\newblock {\em CoRR}, abs/1703.00548, 2017.

\bibitem{DBLP:journals/corr/abs-1810-05749}
Chris Zhang, Mengye Ren, and Raquel Urtasun.
\newblock Graph hypernetworks for neural architecture search.
\newblock {\em CoRR}, abs/1810.05749, 2018.

\bibitem{pmlr-v80-suganuma18a}
Masanori Suganuma, Mete Ozay, and Takayuki Okatani.
\newblock Exploiting the potential of standard convolutional autoencoders for
  image restoration by evolutionary search.
\newblock In Jennifer Dy and Andreas Krause, editors, {\em Proceedings of the
  35th International Conference on Machine Learning}, volume~80 of {\em
  Proceedings of Machine Learning Research}, pages 4771--4780,
  Stockholmsmässan, Stockholm Sweden, 7 2018. PMLR.

\bibitem{7273701}
S.~{Lander} and Y.~{Shang}.
\newblock Evoae -- a new evolutionary method for training autoencoders for deep
  learning networks.
\newblock In {\em 2015 IEEE 39th Annual Computer Software and Applications
  Conference}, volume~2, pages 790--795, 7 2015.

\bibitem{charte2019automating}
Francisco Charte, Antonio~J Rivera, Francisco Mart{\'\i}nez, and Mar{\'\i}a~J
  del Jesus.
\newblock Automating autoencoder architecture configuration: An evolutionary
  approach.
\newblock In {\em International Work-Conference on the Interplay Between
  Natural and Artificial Computation}, pages 339--349. Springer, 2019.

\bibitem{DBLP:journals/corr/abs-1902-08142}
Christian Sciuto, Kaicheng Yu, Martin Jaggi, Claudiu Musat, and Mathieu
  Salzmann.
\newblock Evaluating the search phase of neural architecture search.
\newblock {\em CoRR}, abs/1902.08142, 2019.

\bibitem{10.1007/978-3-319-10590-1_53}
Matthew~D. Zeiler and Rob Fergus.
\newblock Visualizing and understanding convolutional networks.
\newblock In David Fleet, Tomas Pajdla, Bernt Schiele, and Tinne Tuytelaars,
  editors, {\em Computer Vision -- ECCV 2014}, pages 818--833, Cham, 2014.
  Springer International Publishing.

\bibitem{DBLP:journals/corr/SzegedyVISW15}
Christian Szegedy, Vincent Vanhoucke, Sergey Ioffe, Jonathon Shlens, and
  Zbigniew Wojna.
\newblock Rethinking the inception architecture for computer vision.
\newblock {\em CoRR}, abs/1512.00567, 2015.

\bibitem{DBLP:conf/cvpr/HeZRS16}
Kaiming He, Xiangyu Zhang, Shaoqing Ren, and Jian Sun.
\newblock Deep residual learning for image recognition.
\newblock In {\em {CVPR}}, pages 770--778. {IEEE} Computer Society, 2016.

\bibitem{lin2008riemannian}
Tong Lin and Hongbin Zha.
\newblock Riemannian manifold learning.
\newblock {\em IEEE Transactions on Pattern Analysis and Machine Intelligence},
  30(5):796--809, 2008.

\bibitem{Tenenbaum2319}
Joshua~B. Tenenbaum, Vin~de Silva, and John~C. Langford.
\newblock A global geometric framework for nonlinear dimensionality reduction.
\newblock {\em Science}, 290(5500):2319--2323, 2000.

\bibitem{Roweis00nonlineardimensionality}
Sam~T. Roweis and Lawrence~K. Saul.
\newblock Nonlinear dimensionality reduction by locally linear embedding.
\newblock {\em SCIENCE}, 290:2323--2326, 2000.

\bibitem{Belkin:2003:LED:795523.795528}
Mikhail Belkin and Partha Niyogi.
\newblock Laplacian eigenmaps for dimensionality reduction and data
  representation.
\newblock {\em Neural Comput.}, 15(6):1373--1396, June 2003.

\bibitem{Abdi:2010:PCA:3160436.3160440}
Herv{\'e} Abdi and Lynne~J. Williams.
\newblock Principal component analysis.
\newblock {\em WIREs Comput. Stat.}, 2(4):433--459, July 2010.

\bibitem{DBLP:conf/cvpr/TompsonGJLB15}
Jonathan Tompson, Ross Goroshin, Arjun Jain, Yann LeCun, and Christoph Bregler.
\newblock Efficient object localization using convolutional networks.
\newblock In {\em {IEEE} Conference on Computer Vision and Pattern Recognition,
  {CVPR} 2015, Boston, MA, USA, June 7-12, 2015}, pages 648--656. {IEEE}
  Computer Society, 2015.

\bibitem{hajewski-bdl}
Jeff Hajewski and Suely Oliveira.
\newblock A scalable system for neural architecture search.
\newblock In {\em IEEE Computing and Communication Workshop and Conference},
  CCWC'20, 2020.

\bibitem{grpc}
Google.
\newblock grpc.

\bibitem{Varda2008}
Kenton Varda.
\newblock Protocol buffers: Google's data interchange format.
\newblock Technical report, Google, 6 2008.

\bibitem{paszke2017automatic}
Adam Paszke, Sam Gross, Soumith Chintala, Gregory Chanan, Edward Yang, Zachary
  DeVito, Zeming Lin, Alban Desmaison, Luca Antiga, and Adam Lerer.
\newblock Automatic differentiation in {PyTorch}.
\newblock In {\em NIPS Autodiff Workshop}, 2017.

\bibitem{DBLP:journals/jmlr/CoatesNL11}
Adam Coates, Andrew~Y. Ng, and Honglak Lee.
\newblock An analysis of single-layer networks in unsupervised feature
  learning.
\newblock In Geoffrey~J. Gordon, David~B. Dunson, and Miroslav Dud{\'i}k,
  editors, {\em Proceedings of the Fourteenth International Conference on
  Artificial Intelligence and Statistics, {AISTATS} 2011, Fort Lauderdale, USA,
  April 11-13, 2011}, volume~15 of {\em {JMLR} Proceedings}, pages 215--223.
  JMLR.org, 2011.

\bibitem{cifar10}
Alex Krizhevsky, Vinod Nair, and Geoffrey Hinton.
\newblock Cifar-10 (canadian institute for advanced research).

\bibitem{43826}
David~K. Rensin.
\newblock {\em Kubernetes - Scheduling the Future at Cloud Scale}.
\newblock 1005 Gravenstein Highway North Sebastopol, CA 95472, 2015.

\bibitem{Hindman:2011:MPF:1972457.1972488}
Benjamin Hindman, Andy Konwinski, Matei Zaharia, Ali Ghodsi, Anthony~D. Joseph,
  Randy Katz, Scott Shenker, and Ion Stoica.
\newblock Mesos: A platform for fine-grained resource sharing in the data
  center.
\newblock In {\em Proceedings of the 8th USENIX Conference on Networked Systems
  Design and Implementation}, NSDI'11, pages 295--308, Berkeley, CA, USA, 2011.
  USENIX Association.

\end{thebibliography}

\end{document}